\documentclass{article}

\usepackage[preprint]{neurips_2026}

\usepackage[utf8]{inputenc}
\usepackage[T1]{fontenc}
\usepackage{hyperref}
\usepackage{url}
\hypersetup{hidelinks}
\usepackage{booktabs}
\usepackage{amsfonts}
\usepackage{amsmath}
\usepackage{amssymb}
\usepackage{graphicx}
\usepackage{float}
\usepackage{subcaption}
\usepackage{multirow}
\usepackage{microtype}
\usepackage[table]{xcolor}
\usepackage{caption}

\graphicspath{{figures/}}

\title{Drift Calibration in Geometric Eye Tracking Systems}

\author{\textbf{Jiaqi Liu}\textsuperscript{\textdagger,\textdaggerdbl,*} \quad
  \textbf{Zixuan Wang}\textsuperscript{\textdagger} \quad \textbf{Yuhong Zhang} \quad \textbf{Dingkang Liang}\\
  \textbf{Jane Hanqi Li} \quad \textbf{Tzyy-Ping Jung}\textsuperscript{\S} \quad \textbf{Gert Cauwenberghs}\textsuperscript{\S}\\[2pt]
  \normalfont Institute for Neural Computation, University of California San Diego
}

\makeatletter
\renewcommand{\@noticestring}{%
  \noindent\textsuperscript{\textdagger} Co-first authors. \quad
  \textsuperscript{\textdaggerdbl} Project leader. \quad
  \textsuperscript{\S} Corresponding authors.\par
  \noindent\textsuperscript{*} Contact:
  \href{mailto:jil585@ucsd.edu}{\texttt{jil585@ucsd.edu}}.\hfill Preprint.%
}
\makeatother

\begin{document}

\maketitle

\begin{abstract}
Geometric eye trackers can provide the spatial accuracy required for gaze-based interaction and multimodal studies, but their measurements remain sensitive to residual session-specific calibration error. Research on correcting this error is difficult to compare because methods are typically evaluated with different devices, target layouts, and error definitions. We present a calibration-focused dataset containing 163 trials from 12 participants, with separate 18-point fitting and 32-point test grids, and use it to evaluate global, local, and composite correction functions under a common spatial-extrapolation protocol. We further introduce a lightweight neural refiner that combines ranked predictions from complementary calibrators. On this controlled dataset, post-vendor correction reduces the mean angular error from $1.53^\circ$ to $1.03^\circ$ with the best-performing classical composite and to $0.96^\circ$ with the refiner. In a closed-loop gaze task, task performance covaries with residual error across four online correction conditions. These results provide a reproducible data-quality benchmark for using gaze as a behavioral signal in interactive modeling.

\end{abstract}

\begin{figure}[!ht]
\centering
\includegraphics[width=0.750\linewidth]{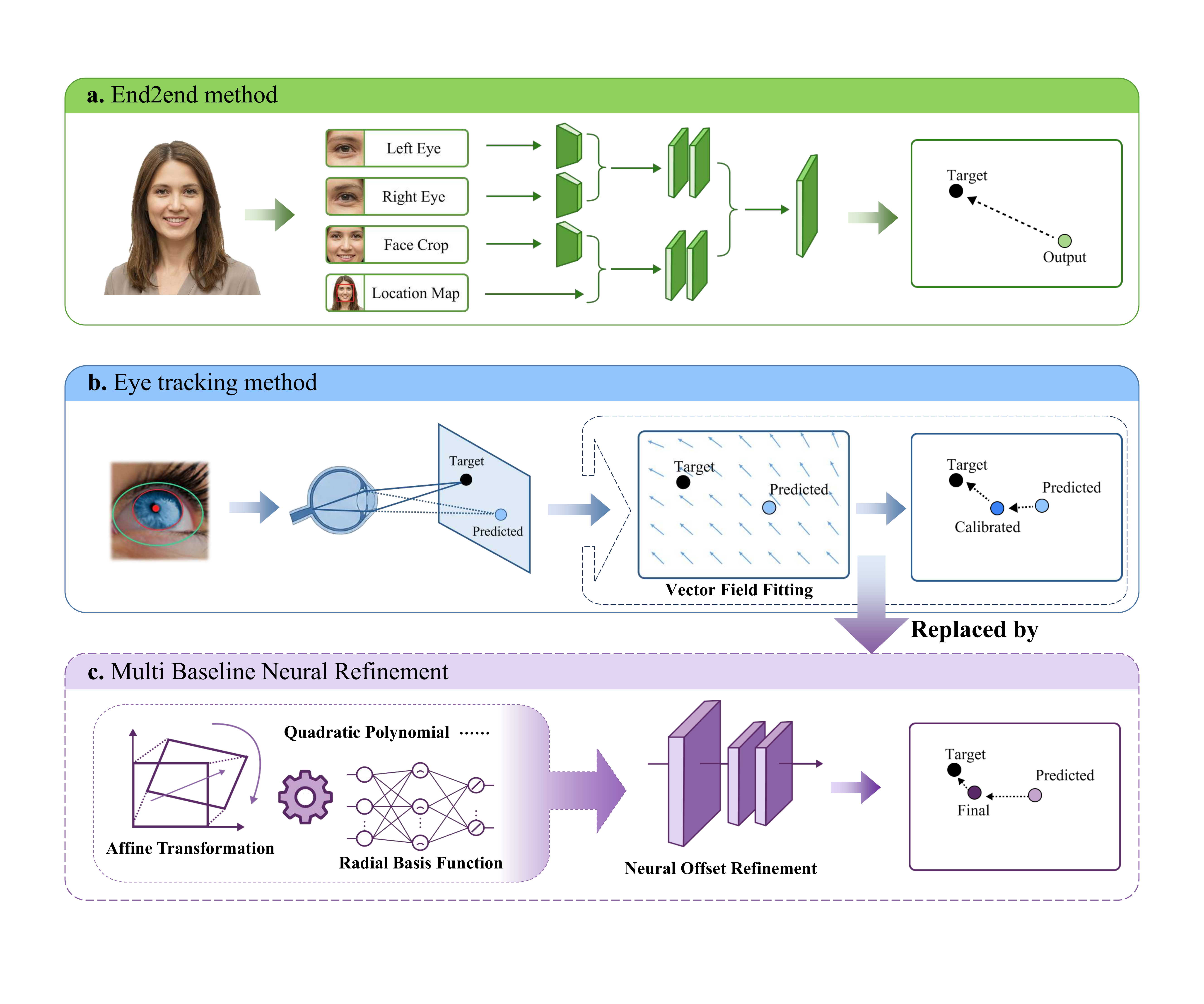}
\caption{End-to-end gaze estimation, geometric calibration, and neural refinement over multiple calibrators.}
\label{fig:taxonomy}
\end{figure}

\section{Introduction}
\label{sec:intro}
Gaze links visual behavior to locations in a scene or interface, making it useful both as a behavioral signal and as an input to interactive systems. It is combined with neural measurements and used in brain--computer interfaces, clinical assessment, and immersive interaction~\citep{Admoni2017, Wang2015, TahriSqalli2023, Wang2020, Emerson2020}. In each setting, the value of a gaze sample depends on assigning it to the intended word, facial feature, object, or target of the interface. A systematic spatial offset can therefore change the behavioral label attached to a sample or the action triggered by a closed-loop system. Gaze data quality is consequently task-relative: accuracy must be judged against the size and spacing of the units used by the downstream analysis or interaction. When recordings are pooled across participants and sessions, calibration differences introduce measurement heterogeneity that a downstream model may otherwise absorb as behavioral variation.

Two broad approaches produce screen-referenced gaze estimates. Appearance-based models learn a direct mapping from eye or face images to gaze~\citep{Krafka2016, Park2019FAZE, Park2020, Balim2023, Wang2023}. Geometric trackers instead infer an optical gaze ray from a near-eye camera and use a short, per-session procedure to map that ray onto the display~\citep{Kassner2014}. These approaches use different sensors and are typically evaluated on different benchmarks, so their reported angular errors are not directly comparable (Figure~\ref{fig:taxonomy}). We focus on a narrower problem: the structured spatial error that remains in a geometric tracker's two-dimensional output after vendor calibration. Headset placement and slippage are established sources of such errors~\citep{Cognolato2018, Niehorster2020}, but our controlled recordings do not isolate a single cause. We therefore treat the residual as a session-specific displacement field estimated from a small set of known targets. In our data set, post-vendor correction reduces the mean angular error from $1.53^\circ$ to $1.03^\circ$ with the strongest classical composite, motivating a controlled comparison of calibration functions.

Calibration research spans several settings that are often conflated. Some methods fit the initial mapping from eye features to gaze using polynomial, spline, radial-basis, Gaussian-process, or regularized regression models~\citep{Cerrolaza2012, Scheel2016, Tripathi2017, Mardanbegi2018, Su2020, Severitt2023}. Others correct an existing tracker's output from explicit validation targets~\citep{Vadillo2015, Lander2016, Padikal2025} or update calibration implicitly from saccades, reading, touch, and interface events~\citep{Huang2019, Philipp2019, Wan2024, Hou2025}. Appearance-model personalization is a separate problem because it adapts an image-to-gaze estimator rather than calibrated two-dimensional output~\citep{Park2019FAZE}. Across these settings, devices, target layouts, error definitions, and evaluation splits differ, leaving unresolved how post-vendor correction families compare when they are fitted and evaluated at disjoint screen locations. We address this question with a common protocol and a lightweight neural refiner that combines predictions from multiple classical calibration functions.

\paragraph{Contributions.} (1) We provide a common empirical basis for post-vendor gaze calibration and connect offline accuracy to closed-loop control. (2) We release an open dataset of 163 trials from 12 participants and systematically compare global, local, and composite corrections. (3) Building on this comparison, we introduce a lightweight neural refiner that combines predictions from seven complementary calibrators and achieves the lowest error in our evaluation ($0.96^\circ$). (4) Across four correction conditions in a closed-loop gaze-control game, a lower residual error corresponds to higher game scores, linking the offline calibration gains to better control performance under the same task conditions.

\section{A Calibration-Oriented Dataset}
\label{sec:dataset}

Datasets for appearance-based gaze estimation usually pair eye or face images with target coordinates and are designed to learn an image-to-gaze mapping. Evaluating post-vendor correction requires a different data structure: tracker-reported gaze at known screen targets, organized by trial, with fitting locations separated from evaluation locations. Retaining complete fixation samples makes it possible to distinguish systematic spatial offset from within-fixation dispersion. We therefore construct a calibration-focused dataset with disjoint fitting and test grids, repeated trials, and full fixation sample clouds. This design allows every correction family in \S\ref{sec:spectrum} to use the same observations and tests spatial generalization rather than recovering calibration points.

\begin{figure}[tb]
\centering
\begin{subfigure}[b]{0.32\linewidth}\centering
  \includegraphics[width=\linewidth]{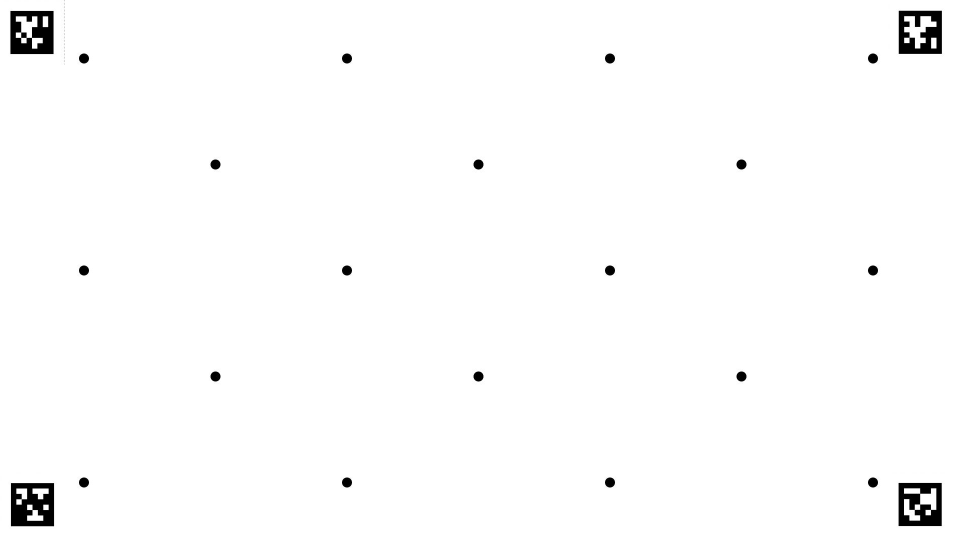}
  \caption{Fitting grid (18 pts)}\end{subfigure}\hfill
\begin{subfigure}[b]{0.32\linewidth}\centering
  \includegraphics[width=\linewidth]{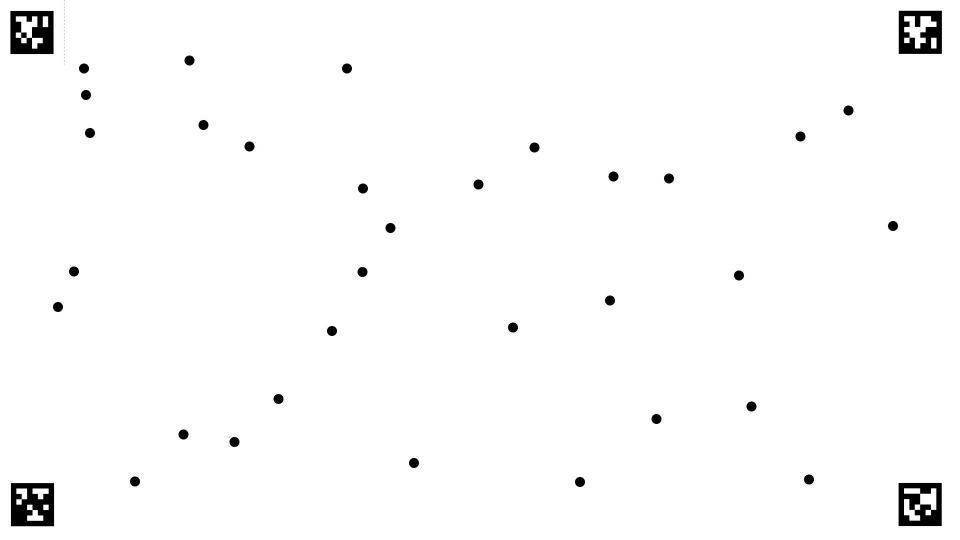}
  \caption{Test grid (32 pts)}\end{subfigure}\hfill
\begin{subfigure}[b]{0.32\linewidth}\centering
  \includegraphics[width=\linewidth]{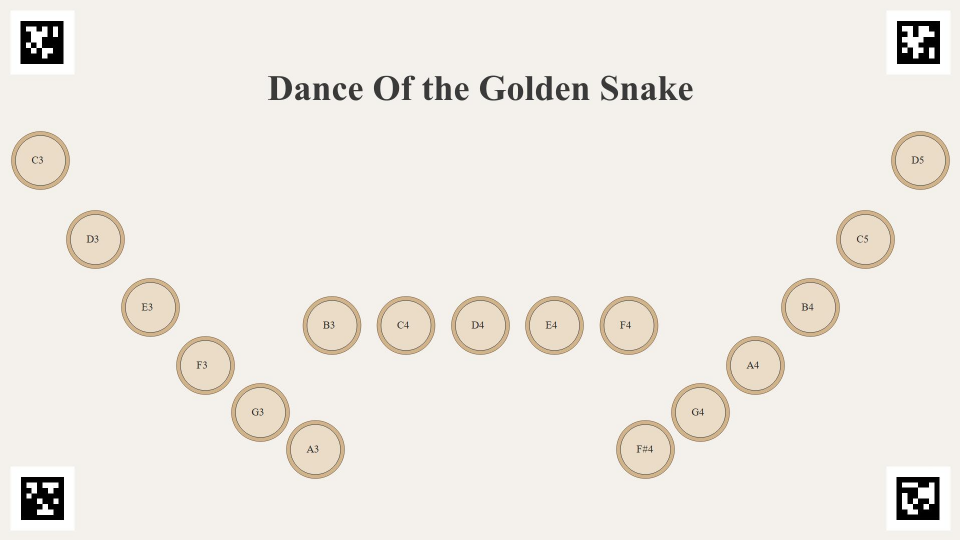}
  \caption{Closed-loop task}\end{subfigure}

\vspace{3pt}
\begin{subfigure}[b]{0.32\linewidth}\centering
  \includegraphics[width=\linewidth]{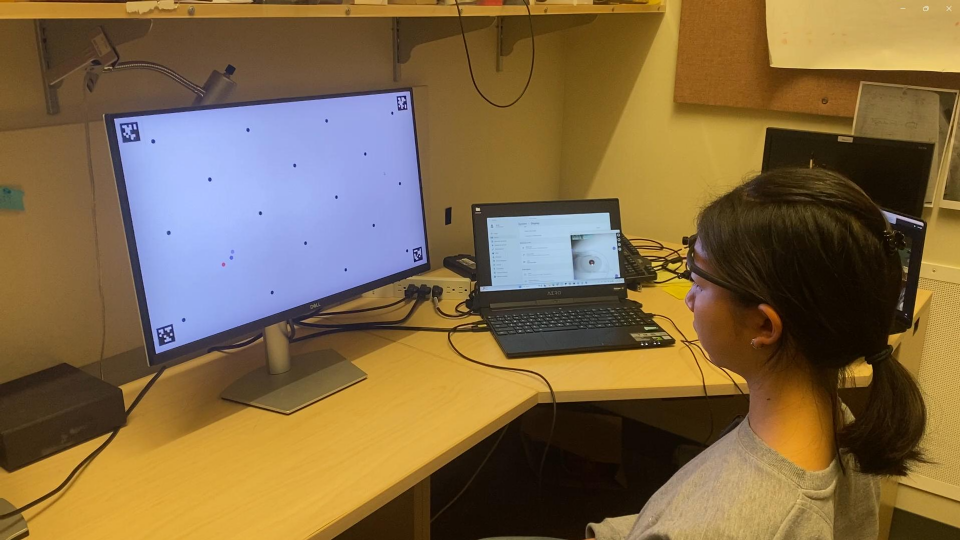}\end{subfigure}\hfill
\begin{subfigure}[b]{0.32\linewidth}\centering
  \includegraphics[width=\linewidth]{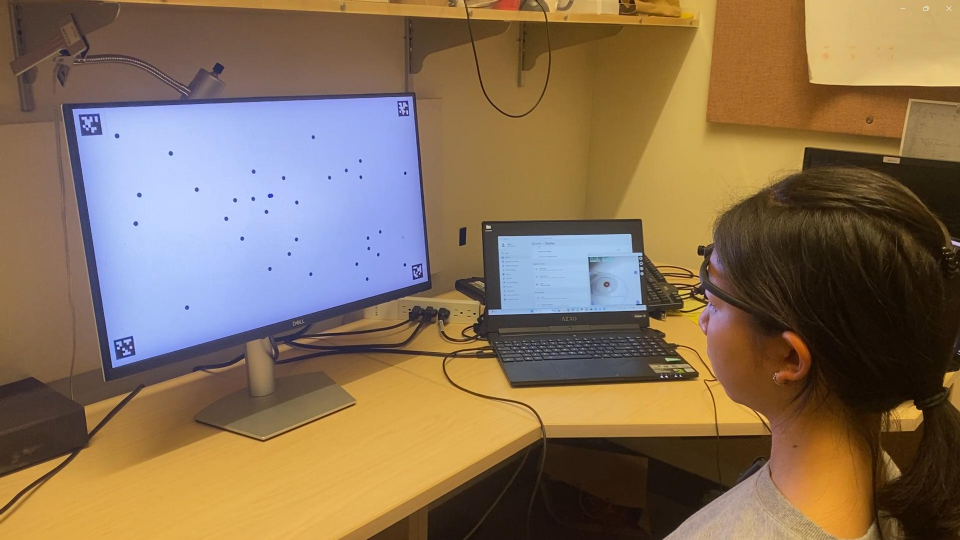}\end{subfigure}\hfill
\begin{subfigure}[b]{0.32\linewidth}\centering
  \includegraphics[width=\linewidth]{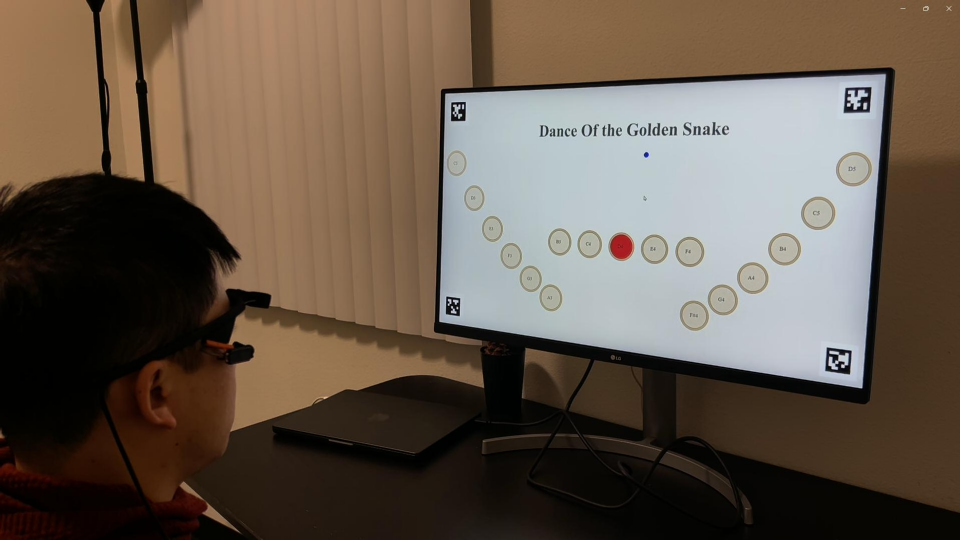}\end{subfigure}
\caption{Acquisition. Top: the three displays. Bottom: matching sessions at a fixed 70\,cm with a head-mounted tracker.}
\label{fig:acquisition}
\end{figure}

Among 12 participants (5 male, 7 female; ages 18--25), we collected 163 trials with a monocular Pupil Labs Core head-mounted tracker~\citep{Kassner2014}, a 24-inch $1920\times1080$ display and a fixed viewing distance of 70 \, cm (Figure~\ref{fig:acquisition}). Each session began with the tracker's standard five-point calibration, so the dataset measures residual error after the vendor procedure rather than initial gaze mapping. Each test then comprised an 18-point fitting grid and a 32-point disjoint test grid. Each target remained visible for 6\,s. We discarded the first 2 \, s because, among more than 2{,}000 recorded target fixations, more than 95\% stabilized within 1.83 \, s (Appendix~\ref{app:fixation}). Separation of the grids prevents scoring at fitted locations and makes the evaluation a test of spatial generalization.

Each fixation retains its full sample cloud. Dispersion measures noise and precision; centroid offset measures drift and accuracy, a distinction that is not available for centroids alone~\citep{Holmqvist2012}. Fields are spatially coherent: a large near-constant component plus a smaller position-dependent one. However, they vary between participants and sessions (Appendix~\ref{app:drift}), which motivates the spectrum. We release processed data, acquisition software, all calibrators, the refiner, and the evaluation harness under a permissive license, with identifiers replaced by opaque codes.

\section{Combining Classical Calibrators with Neural Refinement}
\label{sec:method}

Each trial provides $N$ correspondences between tracker estimates $\mathbf{p}_i$ and known target locations $\mathbf{p}'_i$. The displacements $\boldsymbol{\delta}_i=\mathbf{p}'_i-\mathbf{p}_i$ sample a session-specific drift field at a small set of screen locations. A calibrator predicts this displacement at unseen positions, giving corrected gaze $\mathbf{p}+\mathcal{C}(\mathbf{p})$. The central challenge is therefore spatial extrapolation from sparse, noisy observations. We first compare several classical calibrators under this shared formulation, then combine their predictions with a lightweight neural refiner.

\label{sec:spectrum}

We organize classical calibrators by how strongly they constrain the drift field. Similarity and polynomial models impose a global, smooth correction, while kernel methods such as RBF, thin-plate splines, and Gaussian processes allow more local variation. Composite models first remove a global similarity component and then model the remaining residual with a flexible calibrator. The comparison therefore probes the tradeoff between stable extrapolation from few targets and the capacity to follow position-dependent drift. Each calibrator is fitted independently per trial using only that trial's calibration points. Appendix~\ref{app:families} gives the full definitions and implementation choices.

\begin{figure}[htbp]
\centering
\includegraphics[width=\linewidth]{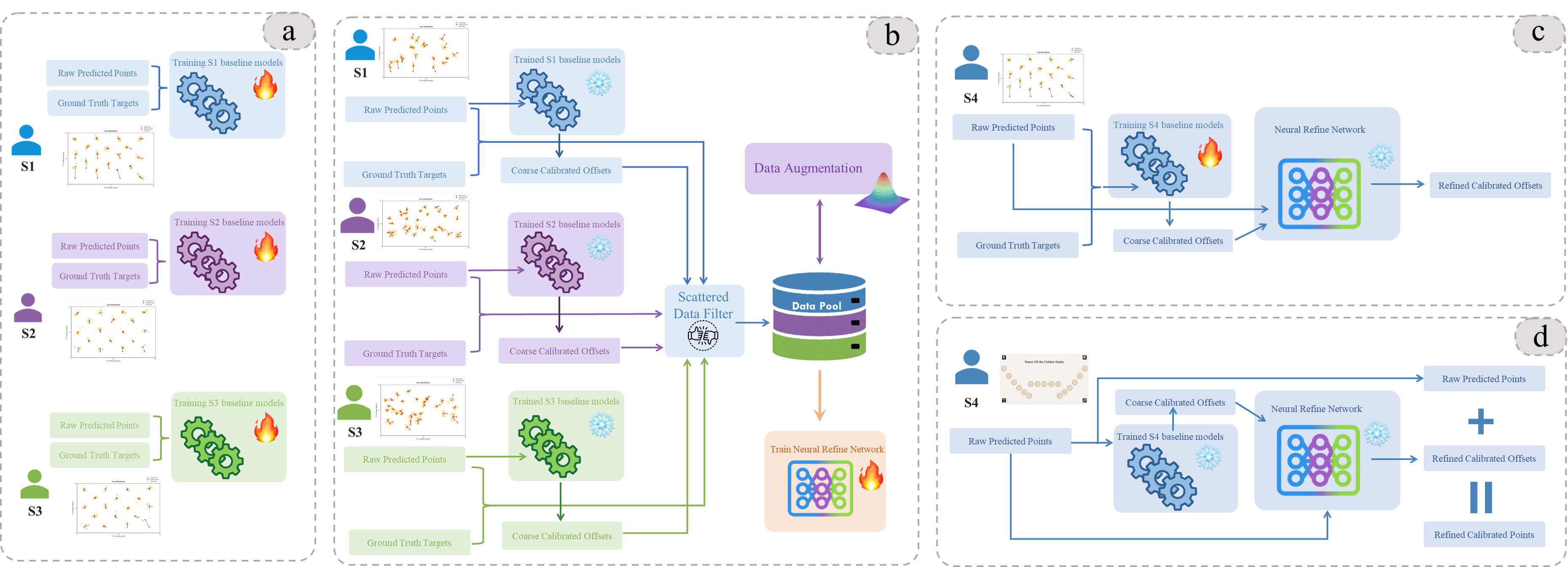}
\caption{Pipeline. (a) Per-trial fitting on 18 points. (b) Ranked hypotheses on the disjoint 32-point grid. (c) Online correction. (d) Closed-loop validation.}
\label{fig:pipeline}
\end{figure}

Because calibrators extrapolate differently, we rank their predictions by residual error on the current trial's fitting grid and place the lowest-residual hypotheses $\boldsymbol{\delta}_{(1)}(\mathbf{p}),\ldots,\boldsymbol{\delta}_{(M)}(\mathbf{p})$ in fixed slots. A small multilayer perceptron combines these hypotheses with the query position $\mathbf{p}$,
\begin{equation}
  \mathbf{p}_{\text{refined}} = \mathbf{p} + f_\theta\big(\mathbf{p},\,
  \boldsymbol{\delta}_{(1)}(\mathbf{p}),\ldots,\boldsymbol{\delta}_{(M)}(\mathbf{p})\big),
  \qquad f_\theta:\mathbb{R}^{2(M+1)}\!\to\mathbb{R}^{2}.
  \label{eq:refiner}
\end{equation}
The position provides spatial context, while the hypothesis set exposes agreement and disagreement among correction families. Ranking uses only calibration points available at deployment. During training, we perturb the hypothesis channels with Gaussian noise while leaving the raw gaze and target unchanged. This encourages robustness to unreliable base calibrators.
\section{Results}
\label{sec:results}

\paragraph{Evaluation protocol.}
For each trial, classical calibrators were fitted on the 18-point calibration grid and evaluated on the spatially disjoint 32-point test grid. Thus, all reported errors measure generalization to unseen screen locations rather than recovery of fitted targets. The neural refiner was evaluated within population, with participants represented in both the training and test partitions; no cross-user generalization is claimed. The final model combines all seven complementary calibration hypotheses, which achieved the lowest error in the hypothesis-count ablation.

\begin{table}[t]
  \centering
  \caption{Gaze error at 32 locations disjoint from the 18 per-trial fitting locations. Classical functions are fitted per trial; the refiner uses the pooled within-population split described in the protocol.}
  \label{tab:main}
  \footnotesize
  \begin{tabular*}{\textwidth}{@{\extracolsep{\fill}}lccccc}
    \toprule
    \multirow{2}{*}{\textbf{Method}} & \multicolumn{3}{c}{\textbf{Mean error}} &
    \multirow{2}{*}{\textbf{Median (px)}} & \multirow{2}{*}{\textbf{SD (px)}}\\
    \cmidrule(lr){2-4}
     & Angle ($^\circ$) & Pixels & $\Delta$ (\%) & & \\
    \midrule
    Raw tracker output~\citep{Kassner2014} & 1.53 & 67.30 & 0.0 & 55.80 & 43.50\\
    \multicolumn{6}{l}{\textit{Single-family corrections}}\\
    Polynomial (order 2)~\citep{Cerrolaza2012} & 1.11 & 48.92 & 27.3 & 37.83 & 39.7\\
    Similarity & 1.15 & 50.50 & 25.0 & 40.60 & 35.1\\
    Thin-plate spline~\citep{Scheel2016} & 1.13 & 49.60 & 26.3 & 40.90 & 34.9\\
    Gaussian process~\citep{Tripathi2017} & 1.10 & 48.30 & 28.2 & 40.80 & 32.6\\
    \midrule
    \multicolumn{6}{l}{\textit{Composite corrections}}\\
    Similarity + GPR & 1.05 & 46.20 & 31.8 & 39.00 & 31.8\\
    Similarity + RBF & 1.03 & 45.50 & 32.3 & 39.00 & 31.9\\
    \midrule
    \multicolumn{6}{l}{\textit{Neural refinement}}\\
    Neural refiner (ours) & \textbf{0.96} & \textbf{42.30} & \textbf{37.2} & \textbf{37.30} & \textbf{25.20}\\
    \bottomrule
  \end{tabular*}
\end{table}

\paragraph{Post-vendor calibration substantially reduces residual error.}
Raw tracker output had a mean error of $1.53^\circ$ (67.3 px), a median of 55.8 px, and a standard deviation of 43.5 px (Table~\ref{tab:main}). Every calibration family reduced mean pixel error by at least 25.0\%, and the strongest single-family methods achieved reductions of 26.3--28.2\%. Composite models performed better: Similarity+GPR reached $1.05^\circ$ (46.2 px), while Similarity+RBF achieved the best classical result of $1.03^\circ$ (45.5 px), 32.3\% below the raw output.

The neural refiner achieved the best overall result: $0.96^\circ$ mean error (42.3 px), a 37.3 px median, and a 25.2 px standard deviation. Relative to raw output, it reduced mean error, median error, and standard deviation by 37.2\%, 33.2\%, and 42.1\%, respectively. It further improved over the best classical composite by 7.0\% in mean error and 21.0\% in standard deviation, demonstrating gains in both accuracy and consistency.

\begin{figure}[t]
  \centering
  \begin{subfigure}[b]{0.345\linewidth}
    \centering
    \includegraphics[width=\linewidth]{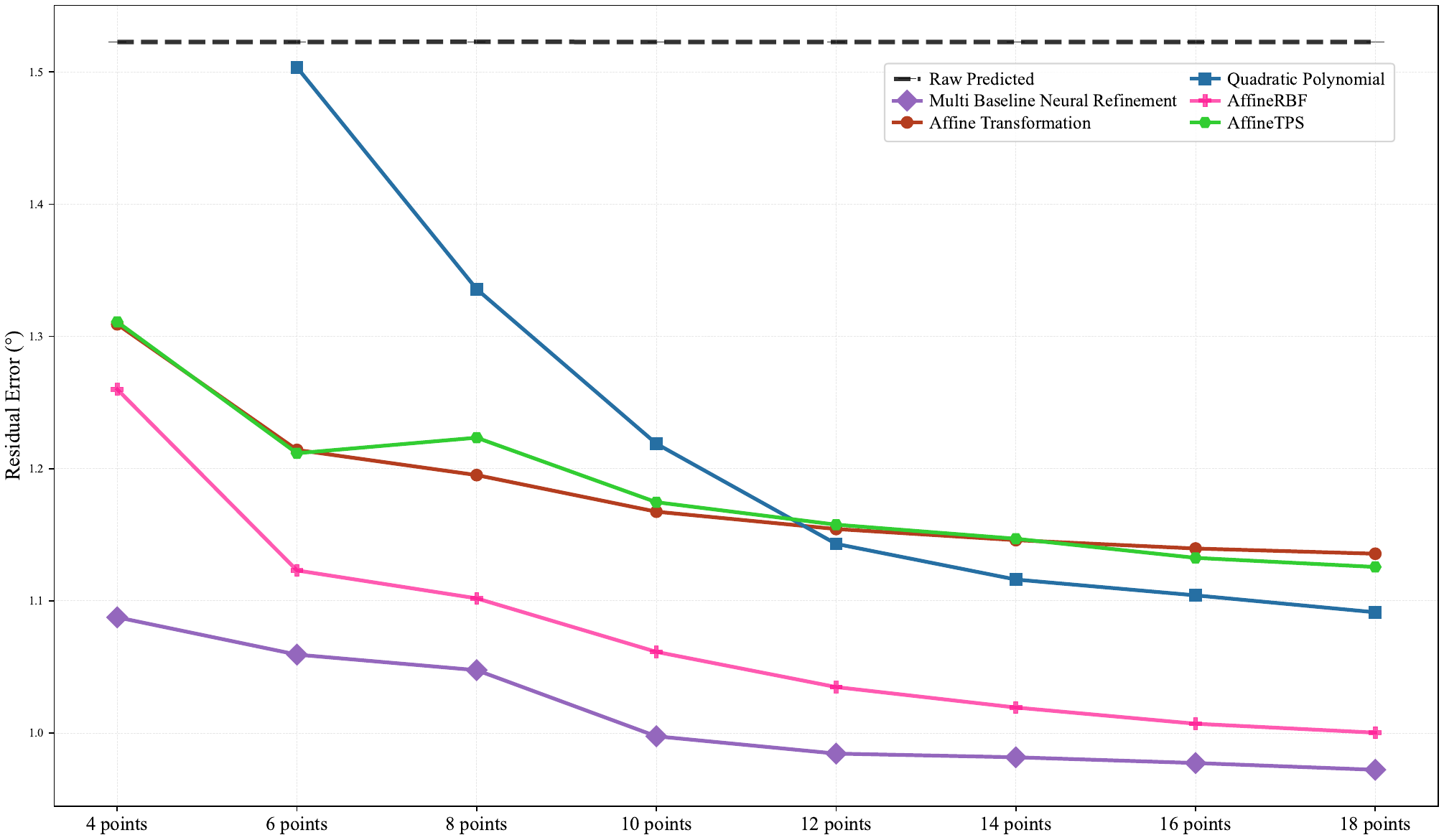}
    \caption{Calibration density}
    \label{fig:density}
  \end{subfigure}\hfill
  \begin{subfigure}[b]{0.325\linewidth}
    \centering
    \includegraphics[width=\linewidth]{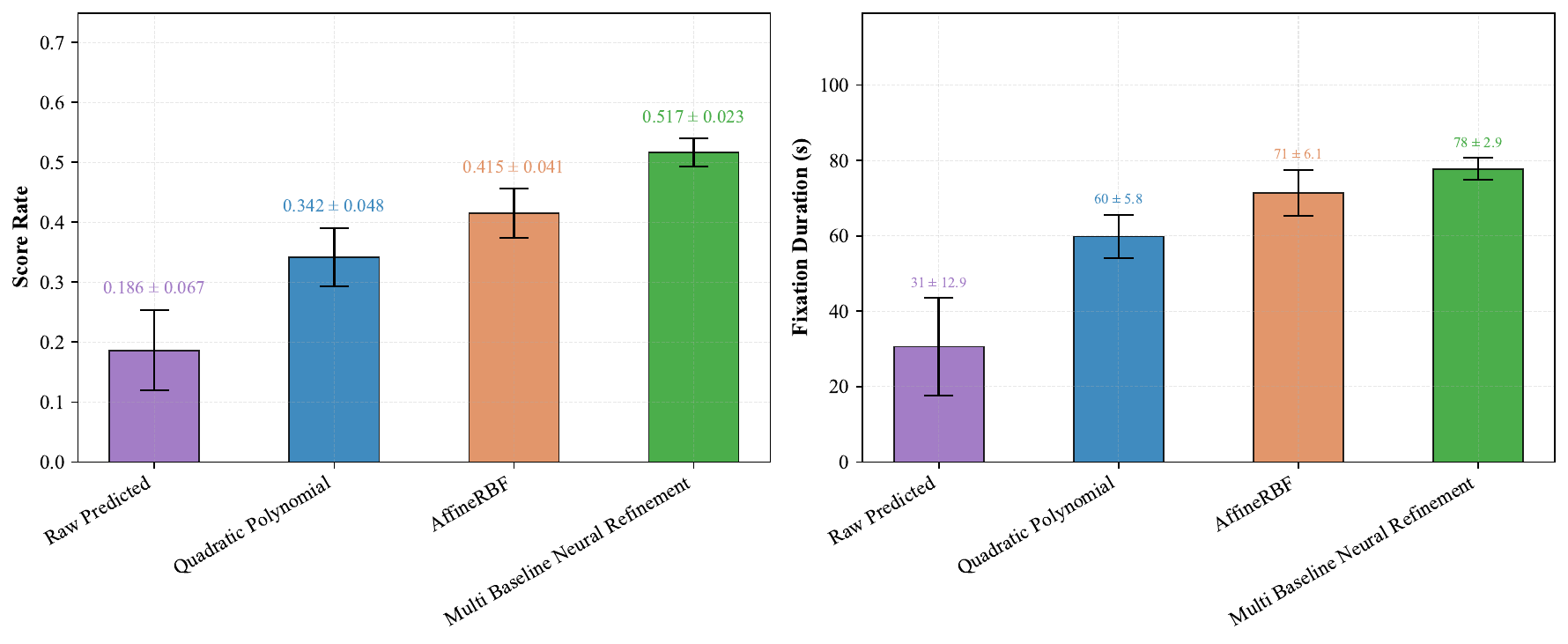}
    \caption{Task score and dwell}
    \label{fig:gamebars}
  \end{subfigure}\hfill
  \begin{subfigure}[b]{0.305\linewidth}
    \centering
    \includegraphics[width=\linewidth]{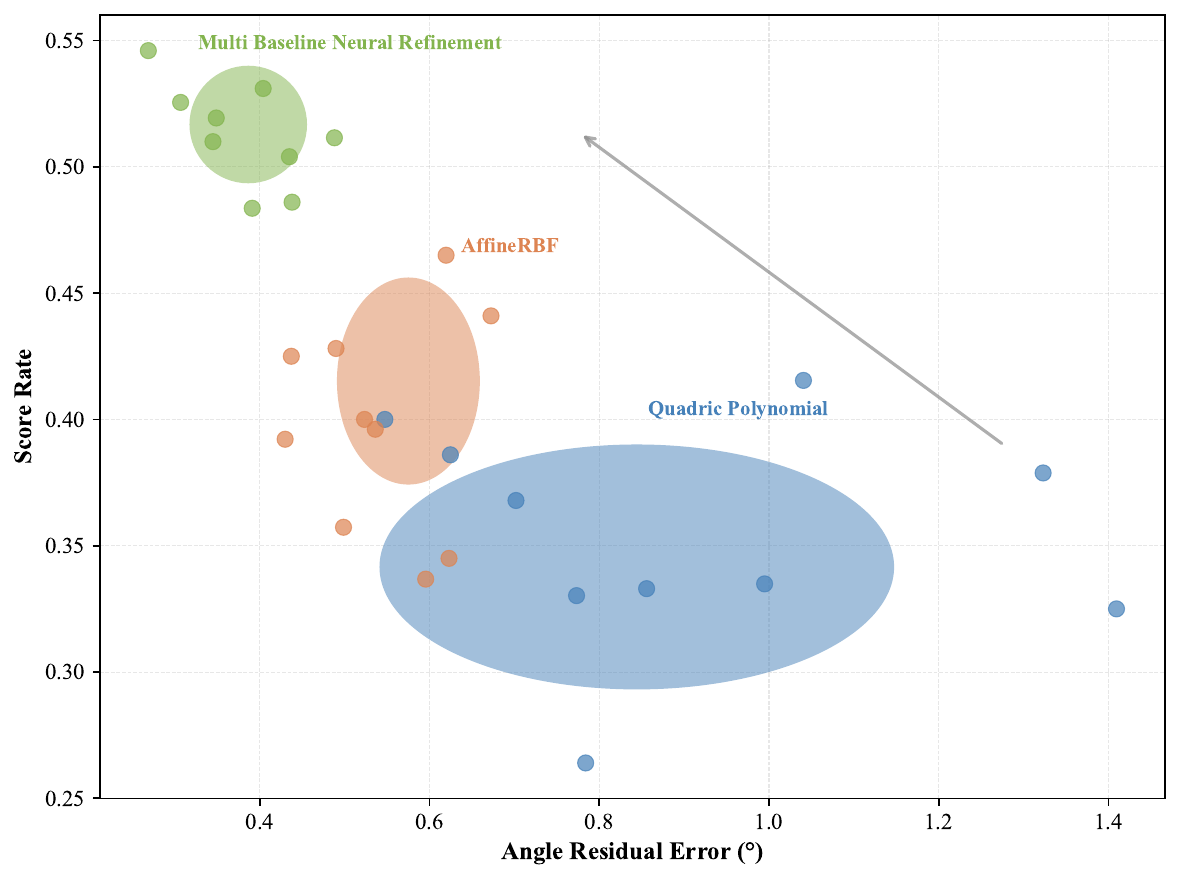}
    \caption{Score vs.\ error}
    \label{fig:score}
  \end{subfigure}
  \caption{Error and closed-loop performance. (a) Error versus calibration density. (b) Score rate and fixation duration in four conditions. (c) Per-trial score versus residual error.}
  \label{fig:results}
\end{figure}

\paragraph{Complementary hypotheses and sparse calibration.}
Error decreased from 44.5 px with one hypothesis to 43.1 px with three and 42.3 px with all seven, giving the full model improvements of 4.9\% and 1.9\%, respectively. The benefit was larger with sparse calibration (Figure~\ref{fig:density}). Using six points, the refiner reached 46.7 px versus 51.4 px for the strongest classical composite, a 9.1\% reduction. Hypothesis perturbation reduced error from 52.4 to 46.7 px with six points, from 47.9 to 44.5 px with 12, and from 42.8 to 42.3 px with 18, corresponding to gains of 10.9\%, 7.1\%, and 1.2\%. Thus, combining and perturbing complementary hypotheses is most valuable when calibration observations are limited.

\paragraph{Closed-loop performance follows residual calibration error.}
Across four online correction conditions, task score increased with calibration quality and was negatively associated with residual angular error (Figures~\ref{fig:gamebars} and~\ref{fig:score}). The common error--performance trend shows that differences near the one-degree range remain relevant to real-time interaction. Together, the full-grid, sparse-calibration, ablation, and within-subject closed-loop results consistently support complementary post-vendor correction under the controlled single-device setup.

\section{Conclusion}
\label{sec:conclusion}

We presented a calibration-focused benchmark and a lightweight neural refiner for residual error after vendor calibration. Across 163 trials from 12 participants, the best classical composite reduced mean error from $1.53^\circ$ (67.3 px) to $1.03^\circ$ (45.5 px), while the seven-hypothesis refiner reached $0.96^\circ$ (42.3 px). This represents a 37.2\% reduction from raw output and a further 7.0\% improvement over the best classical method; error variability fell by 42.1\% and 21.0\%, respectively.

The refiner was particularly effective with sparse calibration, reducing six-point error by 9.1\% over the strongest classical composite, while perturbation contributed a 10.9\% gain. Better calibration also coincided with higher closed-loop task performance. These results show that post-vendor drift is structured, correctable, and relevant to real-time gaze use. Although evaluation was limited to one device and a controlled within-population setting, the released benchmark provides a reproducible basis for future cross-user and cross-device validation.

\bibliographystyle{plainnat}
\bibliography{ref}

\newpage
\appendix
\section{Broader Impact}
\label{app:broader-impact}

Errors in gaze localization can assign a fixation to the wrong word, object, facial feature, or interface control. The dataset and disjoint-grid protocol provide a common basis for comparing correction methods in gaze-based neuroscience, clinical research, assistive systems, and immersive interaction. When gaze is aligned with neural or physiological signals, reducing systematic spatial error may also reduce incorrect event labels in the paired data. More accurate tracking also carries risks. Gaze traces can reveal health-related and biometric information~\citep{TahriSqalli2023, Makowski2020}, and improved accuracy can increase the capacity for surveillance or profiling when data are collected without meaningful consent. Our dataset includes 12 young adults recorded with one device under controlled laboratory conditions. The reported gains require validation before they are applied to other populations, devices, or deployment settings. We replace participant identifiers with opaque codes. Downstream users should nevertheless minimize retained data, follow consent and access constraints, and report performance across devices and participant groups.

\section{Calibration Function Families}
\label{app:families}

This appendix defines the calibration families used in our comparison. A session provides $N$ correspondences $\{(\mathbf{p}_i,\mathbf{p}'_i)\}_{i=1}^N$ between raw estimates $\mathbf{p}_i=(x_i,y_i)^\top$ and known targets $\mathbf{p}'_i$. We define drift as $\boldsymbol{\delta}(\mathbf{p})=\mathbf{p}'-\mathbf{p}$ and use a calibrator $\mathcal{C}:\mathbb{R}^2\to\mathbb{R}^2$ to predict this displacement at unseen positions. We use $\bar{\mathbf{p}}=\frac{1}{N}\sum_i\mathbf{p}_i$ and $\bar{\mathbf{p}}'=\frac{1}{N}\sum_i\mathbf{p}'_i$ for centroids and $\tilde{\mathbf{p}}_i=\mathbf{p}_i-\bar{\mathbf{p}}$, $\tilde{\mathbf{p}}'_i=\mathbf{p}'_i-\bar{\mathbf{p}}'$ for centered coordinates.

\subsection{Similarity transform (Procrustes)}
\label{app:sim}

The similarity family admits isotropic scale, rotation, and translation, giving four free parameters:
\begin{equation}
  \mathcal{C}_{\text{sim}}(\mathbf{p}) = s\mathbf{R}\mathbf{p}+\mathbf{t}-\mathbf{p},
  \qquad
  \mathbf{R}=\begin{pmatrix}\cos\theta & -\sin\theta\\ \sin\theta & \cos\theta\end{pmatrix}\in SO(2).
  \label{eq:sim-form}
\end{equation}
We obtain the parameters by solving the orthogonal Procrustes problem $\min_{s,\mathbf{R},\mathbf{t}}\sum_i\|\mathbf{p}'_i-(s\mathbf{R}\mathbf{p}_i+\mathbf{t})\|_2^2$. Translation is determined by the centroids. Rotation and scale are computed from the singular value decomposition of the cross-covariance $\mathbf{H}=\sum_i\tilde{\mathbf{p}}_i\tilde{\mathbf{p}}_i'^\top=\mathbf{U}\boldsymbol{\Sigma}\mathbf{V}^\top$:
\begin{equation}
  \mathbf{R}^\star=\mathbf{V}\,\mathrm{diag}(1,\det(\mathbf{V}\mathbf{U}^\top))\,\mathbf{U}^\top,
  \qquad
  s^\star=\frac{\mathrm{tr}(\boldsymbol{\Sigma}\,\mathrm{diag}(1,\det(\mathbf{V}\mathbf{U}^\top)))}{\sum_i\|\tilde{\mathbf{p}}_i\|_2^2},
  \qquad
  \mathbf{t}^\star=\bar{\mathbf{p}}'-s^\star\mathbf{R}^\star\bar{\mathbf{p}}.
  \label{eq:procrustes}
\end{equation}
The factor $\mathrm{diag}(1,\det(\cdot))$ excludes reflections and constrains $\mathbf{R}^\star$ to $SO(2)$. Without it, a degenerate calibration cloud can yield a reflected fit. Computing the centroids and cross-covariance costs $O(N)$; the remaining singular value decomposition is $2\times2$. The solution is defined when the centered source configuration has nonzero total squared norm.

\subsection{Full affine transform}
\label{app:affine}

Relaxing $s\mathbf{R}$ to an arbitrary $\mathbf{A}\in\mathbb{R}^{2\times2}$ admits anisotropic scale and shear, for six parameters: $\mathcal{C}_{\text{aff}}(\mathbf{p})=\mathbf{A}\mathbf{p}+\mathbf{t}-\mathbf{p}$. Writing $\boldsymbol{\Theta}=[\mathbf{A}\;\mathbf{t}]^\top\in\mathbb{R}^{3\times2}$ and stacking homogeneous inputs $\mathbf{X}\in\mathbb{R}^{N\times3}$ with rows $(x_i,y_i,1)$, the least-squares solution is the normal equation
\begin{equation}
  \boldsymbol{\Theta}^\star=(\mathbf{X}^\top\mathbf{X}+\lambda\mathbf{I})^{-1}\mathbf{X}^\top\mathbf{Y},
  \qquad \mathbf{Y}\in\mathbb{R}^{N\times2}\ \text{with rows}\ \mathbf{p}_i'^\top.
  \label{eq:affine-ls}
\end{equation}
This model accommodates anisotropic scale and shear, but its additional degrees of freedom can increase estimation variance when $N$ is small.

\subsection{Polynomial regression}
\label{app:poly}

For order $n$, expand each coordinate into the monomial basis of total degree at most $n$,
\begin{equation}
  \boldsymbol{\psi}_n(\mathbf{p}) = \big(x^a y^b\big)_{0\le a+b\le n} \in \mathbb{R}^{d_n},
  \qquad d_n=\tfrac{(n+1)(n+2)}{2},
  \label{eq:polybasis}
\end{equation}
so $d_2=6$, $d_3=10$, $d_4=15$. We regress the drift components rather than the absolute positions, fitting independent ridge models per output axis:
\begin{equation}
  \mathbf{w}^\star_c=\arg\min_{\mathbf{w}}\ \sum_{i=1}^{N}\big(\boldsymbol{\psi}_n(\mathbf{p}_i)^\top\mathbf{w}-\delta_{c}(\mathbf{p}_i)\big)^2+\lambda\|\mathbf{w}\|_2^2,
  \qquad c\in\{x,y\},
  \label{eq:polyridge}
\end{equation}
with closed form $\mathbf{w}^\star_c=(\boldsymbol{\Psi}^\top\boldsymbol{\Psi}+\lambda\mathbf{I})^{-1}\boldsymbol{\Psi}^\top\boldsymbol{\delta}_c$. Because the model predicts drift, $\mathbf{w}=\mathbf{0}$ corresponds to applying no correction rather than predicting the screen center. With $N=18$ and $d_4=15$, the order-4 fit has only three more observations than coefficients per output, which can increase estimation variance.

\subsection{Radial basis function interpolation}
\label{app:rbf}

The RBF interpolation represents the drift field as a weighted sum of kernels centered on the calibration targets $\{\mathbf{c}_j\}$:
\begin{equation}
  \mathcal{C}_{\text{rbf}}(\mathbf{p})=\sum_{j=1}^{N}\mathbf{w}_j\,\phi(\|\mathbf{p}-\mathbf{c}_j\|_2),
  \qquad
  \boldsymbol{\Phi}\mathbf{W}=\mathbf{D},\quad \Phi_{ij}=\phi(\|\mathbf{c}_i-\mathbf{c}_j\|_2),
  \label{eq:rbf-full}
\end{equation}
where $\mathbf{W}\in\mathbb{R}^{N\times2}$ contains the weights and $\mathbf{D}\in\mathbb{R}^{N\times2}$ contains the observed drifts. Equation~\ref{eq:rbf-full} describes an unsmoothed fit that interpolates the calibration points. Its zero fitting residual cannot rank unsmoothed RBF fits under the criterion in \S\ref{sec:method}; smoothed variants need not interpolate. We use the multiquadric
\begin{equation}
  \phi(r)=\sqrt{r^2+\epsilon^2},
  \label{eq:multiquadric}
\end{equation}
with shape parameter $\epsilon$. The limit $\epsilon\to0$ gives the conical kernel $\phi(r)=r$, while larger $\epsilon$ produces a flatter kernel over the sampled distance range. SciPy infers $\epsilon$ from node spacing; we evaluate smoothing values 0, 1, and 2. Flat systems can be poorly conditioned, and extrapolation also depends on target geometry. We therefore evaluate RBF variants away from the calibration grid.

\subsection{Thin-plate spline}
\label{app:tps}

The thin-plate spline uses the kernel associated with the bending-energy penalty $\int(f_{xx}^2+2f_{xy}^2+f_{yy}^2)\,dx\,dy$. With regularization $\lambda$, its kernel and augmented system are
\begin{equation}
  \phi_{\text{tps}}(r)=r^2\log r,
  \qquad
  \begin{pmatrix}\boldsymbol{\Phi}+\lambda\mathbf{I} & \mathbf{P}\\ \mathbf{P}^\top & \mathbf{0}\end{pmatrix}
  \begin{pmatrix}\mathbf{W}\\ \mathbf{A}\end{pmatrix}
  =\begin{pmatrix}\mathbf{D}\\ \mathbf{0}\end{pmatrix},
  \label{eq:tps}
\end{equation}
where $\mathbf{P}\in\mathbb{R}^{N\times3}$ has rows $(1,x_j,y_j)$ and $\mathbf{A}\in\mathbb{R}^{3\times2}$ contains the affine term. The side condition $\mathbf{P}^\top\mathbf{W}=\mathbf{0}$ separates the affine and non-affine components and reproduces an affine mapping for affine observations. Both terms contribute outside the calibration hull, so the side condition does not constrain extrapolation. The custom TPS implementation uses $\lambda=10^{-3}$.

\subsection{Gaussian process regression}
\label{app:gpr}

Placing a zero-mean GP prior on each drift component with covariance $k$ and noise $\sigma_n^2$ gives the posterior mean
\begin{equation}
  \mathcal{C}_{\text{gpr}}(\mathbf{p})=\mathbf{k}_*^\top(\mathbf{K}+\sigma_n^2\mathbf{I})^{-1}\mathbf{D},
  \qquad
  [\mathbf{k}_*]_j=k(\mathbf{p},\mathbf{c}_j),\quad K_{ij}=k(\mathbf{c}_i,\mathbf{c}_j),
  \label{eq:gpr}
\end{equation}
with the squared-exponential kernel $k(\mathbf{u},\mathbf{v})=\sigma_f^2\exp(-\|\mathbf{u}-\mathbf{v}\|_2^2/2\ell^2)$. GPR differs from exact RBF interpolation in two relevant respects. First, the ridge $\sigma_n^2\mathbf{I}$ permits a nonzero training residual and can reduce sensitivity to fixational noise. Second, the posterior variance $\sigma^2(\mathbf{p})=k(\mathbf{p},\mathbf{p})-\mathbf{k}_*^\top(\mathbf{K}+\sigma_n^2\mathbf{I})^{-1}\mathbf{k}_*$ provides position-dependent uncertainty, which generally increases when a query is weakly supported by the calibration points. We do not use this variance; supplying it to the refiner as another input channel remains untested.

\subsection{Piecewise affine over a Delaunay triangulation}
\label{app:pwa}

Triangulating the calibration targets and fitting an independent affine map per triangle gives a field that is continuous but not differentiable across edges. For a query $\mathbf{p}$ inside the triangle with vertices $(\mathbf{c}_a,\mathbf{c}_b,\mathbf{c}_c)$, barycentric coordinates $(\alpha,\beta,\gamma)$ with $\alpha+\beta+\gamma=1$ give
\begin{equation}
  \mathcal{C}_{\text{pwa}}(\mathbf{p})=\alpha\,\boldsymbol{\delta}(\mathbf{c}_a)+\beta\,\boldsymbol{\delta}(\mathbf{c}_b)+\gamma\,\boldsymbol{\delta}(\mathbf{c}_c).
  \label{eq:pwa}
\end{equation}
Piecewise-affine interpolation is applied only inside the convex hull of the calibration points. For an outside-hull query, the implementation returns the target associated with the nearest control point.

\subsection{Composite calibrators}
\label{app:composite}

For similarity$+$RBF and similarity$+$GPR, the second model fits the residual of the similarity prediction at the raw gaze coordinates. With a similarity displacement $g$ and a residual fitter $h$,
\begin{equation}
  \mathcal{C}_{\text{comp}}(\mathbf{p})=\underbrace{g(\mathbf{p})}_{\text{global}}+\underbrace{h(\mathbf{p})}_{\text{residual}},
  \qquad
  h\ \text{fitted to}\ \{(\mathbf{p}_i,\ \mathbf{p}'_i-\mathbf{p}_i-g(\mathbf{p}_i))\}_{i=1}^N.
  \label{eq:composite-full}
\end{equation}
Composite calibration first fits a global anchor and then models its residual. On our dataset, similarity$+$RBF has lower error than similarity and the other reported single-family baselines in Table~\ref{tab:main}. This result is consistent with residual modeling but does not establish why the composite performs better.

\section{Observed Drift Fields}
\label{app:drift}

\begin{figure}[h]
\centering
\includegraphics[width=\linewidth]{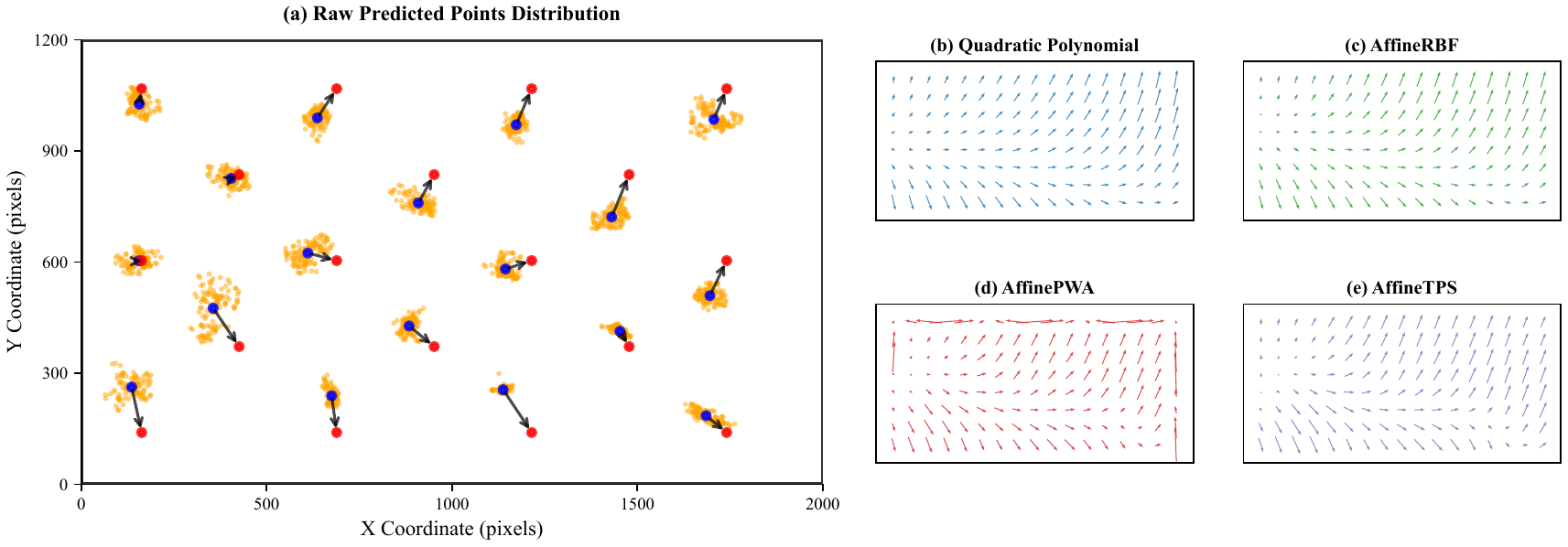}
\caption{Structured drift in one trial. (a) Raw estimates (blue) are displaced from targets (red); yellow marks the fixation cloud. (b--e) Four recovered fields extrapolate differently beyond the calibration points.}
\label{fig:drift}
\end{figure}

\section{Reported Accuracy in Calibration-Related Literature}
\label{app:review}

Table~\ref{tab:review} reports accuracies from two decades of calibration research. Studies differ in device, viewing distance, target count, output dimensionality, external tracking, and error definition; their values should not be compared directly. Section~\ref{sec:dataset} instead evaluates correction methods under a common protocol.

\begin{table}[h]
\centering
\small
\caption{Representative calibration-related results. Values use each study's apparatus and metric and are not comparable across rows. A dash marks an omitted or unavailable value.}
\label{tab:review}
\begin{tabular}{llllr}
\toprule
Year & Author & Device & Method & Accuracy ($^\circ$)\\
\midrule
2012 & \citeauthor{Cerrolaza2012} & Remote VOG & Polynomial families & --\\
2015 & \citeauthor{Vadillo2015}   & EyeLink II & Linear transformation & 0.80$^{*}$\\
2016 & \citeauthor{Scheel2016}    & Pupil Labs Core & RBF + thin-plate spline & 0.85\\
2016 & \citeauthor{Lander2016}    & Pupil Labs (monocular) & 3-point recalibration & --\\
2017 & \citeauthor{Tripathi2017}  & VR HMD & 9-point GPR / polynomial & 1.25 / 1.22\\
2018 & \citeauthor{Mardanbegi2018}& Head-mounted & Polynomial order study & --\\
2019 & \citeauthor{Ehinger2019}   & EyeLink 1000 & Device validation & 0.57\\
2019 & \citeauthor{Huang2019}     & Tobii & Implicit (saccade-based) & --\\
2020 & \citeauthor{Su2020}        & HMGT & Locally polynomial & 1.09\\
2022 & \citeauthor{Yangyang2022}  & Tobii 4C & Smooth pursuit + neural & 0.40\\
2024 & \citeauthor{Wan2024}       & Pupil Labs & Quadratic + RANSAC & 1.13\\
2025 & \citeauthor{Hou2025}       & Meta Quest Pro & Explicit / implicit RLS & 1.19 / 1.94\\
2025 & \citeauthor{Hu2025}        & EOG & De-drift (simulation / real) & 0.90 / 1.03\\
\bottomrule
\multicolumn{5}{l}{\small $^{*}$last-fixation error after correction in the real-data experiment.}
\end{tabular}
\end{table}

\section{Acquisition Protocol}
\label{app:protocol}

\paragraph{Session structure.} Each session contains a fitting phase (18-point grid) and a test phase (32-point grid) at disjoint screen positions, with an optional closed-loop task. Targets illuminate sequentially in random order for 6\,s each. We discard the first 2\,s and record gaze during the remaining 4\,s.

\paragraph{Recording conditions.} Fixed seat and screen positions kept the participant's right eye 70\,cm from the display; seat height and screen angle were adjusted to keep the eye-to-screen axis normal. Participants did not touch the headset after calibration. Small head movements were permitted unless they moved the display outside the world camera's field of view. Illumination was constant, the room was quiet, and participants performed no secondary task.

\paragraph{Closed-loop task.} Seventeen circles spanning C3--D5 appear on the display. After a 3\,s preparation period, a slowed (0.3$\times$) segment of \emph{Dance of the Golden Snake} plays. Each note illuminates its circle for the participant to fixate. Scoring excludes the first 0.7\,s after illumination, then increases with fixation duration inside the circle. Conditions differ only in the calibration function applied to the live gaze stream.

\section{Fixation Acquisition Time}
\label{app:fixation}

\begin{figure}[h]
\centering
\begin{subfigure}[b]{0.28\linewidth}\centering
  \includegraphics[width=\linewidth]{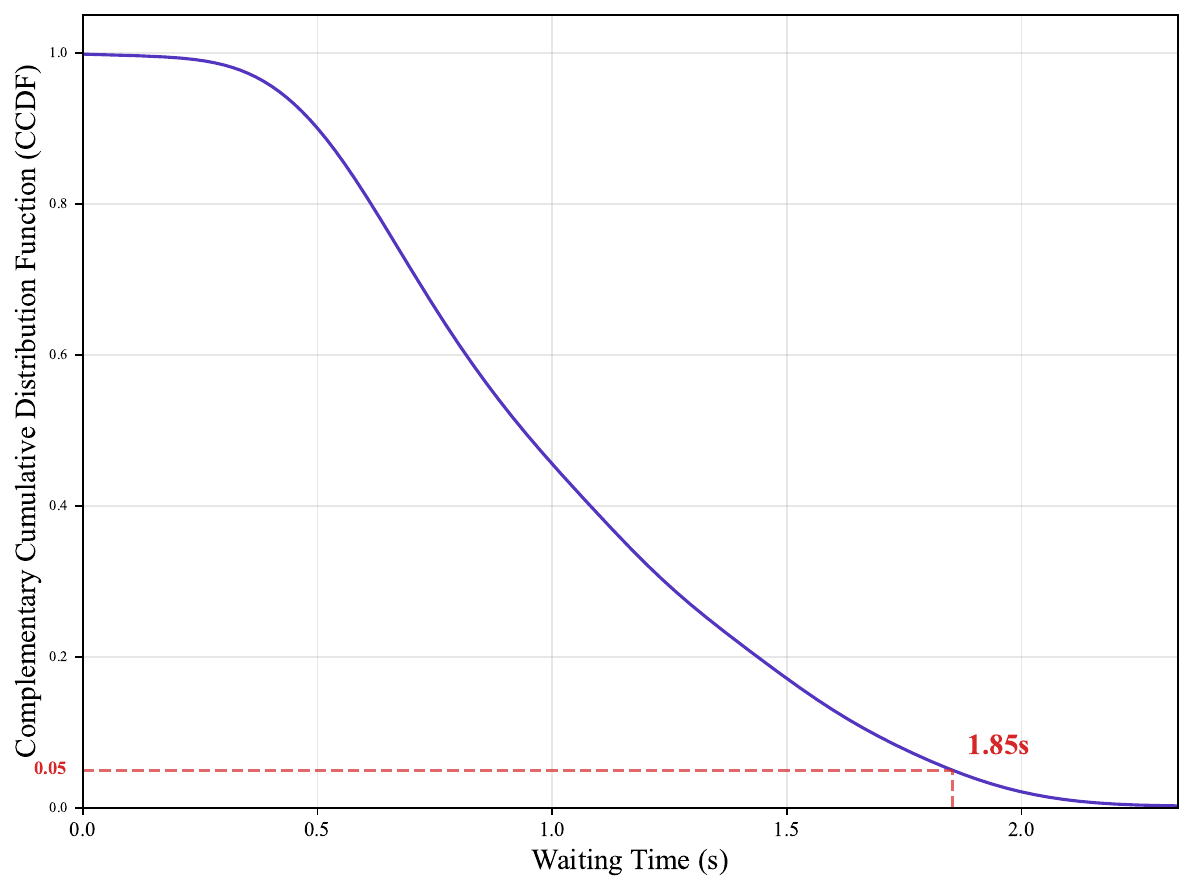}
  \caption*{(a) CCDF of acquisition time}\end{subfigure}\hfill
\begin{subfigure}[b]{0.47\linewidth}\centering
  \includegraphics[width=\linewidth]{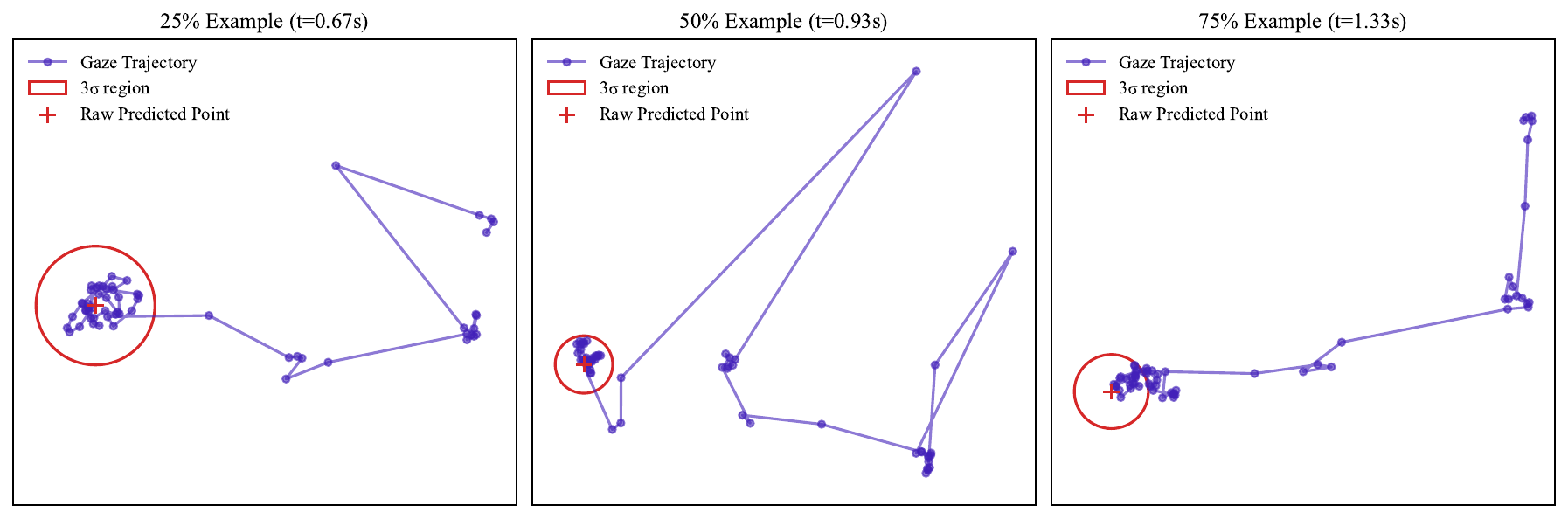}
  \caption*{(b) Trajectories at the 25th, 50th and 75th percentile}\end{subfigure}
\caption{Fixation acquisition relative to the 2\,s discard window.}
\label{fig:ccdf}
\end{figure}

For each target, we compute the mean raw fixation position and its standard deviation $\sigma$ over the 4\,s window. We define a stable region of radius $3\sigma$ around the mean and define acquisition time as the time at which gaze enters and remains in this region. Across more than 2{,}000 collected points, the complementary cumulative distribution shows that over 95\% of targets are acquired within 1.83\,s. This result supports the 2\,s discard threshold (Figure~\ref{fig:ccdf}).

\section{Refiner Ablations}
\label{app:ablation}

Table~\ref{tab:ablation} evaluates hypothesis perturbation and the number of ranked inputs to the refiner.

\begin{table}[H]
\centering
\caption{Left: hypothesis perturbation during training, by calibration density. Right: number of ranked hypotheses supplied to the refiner.}
\label{tab:ablation}
\small
\begin{minipage}[t]{0.55\linewidth}\centering
\begin{tabular}{lccc}
\toprule
\textbf{Training} & \textbf{6 pts} & \textbf{12 pts} & \textbf{18 pts}\\
\midrule
Clean hypotheses & 52.4 & 47.9 & 42.8\\
\rowcolor{gray!12}
Perturbed (ours) & \textbf{46.7} & \textbf{44.5} & \textbf{42.3}\\
\midrule
Gain & 5.7 & 3.4 & 0.5\\
\bottomrule
\end{tabular}
\end{minipage}\hfill
\begin{minipage}[t]{0.42\linewidth}\centering
\begin{tabular}{lc}
\toprule
\textbf{Hypotheses} & \textbf{Error (px)}\\
\midrule
Top 1 & 44.5\\
Top 3 & 43.1\\
\rowcolor{gray!12}
All 7 & \textbf{42.3}\\
\bottomrule
\end{tabular}
\end{minipage}
\end{table}

\end{document}